\documentclass[10pt]{article}
\usepackage[utf8]{inputenc}
\usepackage[T1]{fontenc}
\usepackage{hyperref}
\hypersetup{colorlinks=true, linkcolor=blue, filecolor=magenta, urlcolor=cyan,}
\usepackage{amsmath}
\usepackage{amsfonts}
\usepackage{amssymb}
\usepackage[version=4]{mhchem}
\usepackage{stmaryrd}
\usepackage{bbold}
\usepackage{graphicx}
\usepackage[export]{adjustbox}
\usepackage{underscore}
\usepackage{multirow}

\title{Intelligent OLSR Routing Protocol Optimization for VANETs }

\author{Jamal Toutouh, José García-Nieto, and Enrique Alba}

\begin{document}
\maketitle
\begin{center}
\begin{tabular}{|l|}
\hline
Paper published in the IEEE Transactions on Vehicular Technology journal \\
DOI: 10.1109/TVT.2012.2188552 - https://doi.org/10.1109/TVT.2012.2188552 \\
\end{tabular}
\end{center}

\begin{abstract}
Recent advances in wireless technologies have given rise to the emergence of vehicular ad hoc networks (VANETs). In such networks, the limited coverage of WiFi and the high mobility of the nodes generate frequent topology changes and network fragmentations. For these reasons, and taking into account that there is no central manager entity, routing packets through the network is a challenging task. Therefore, offering an efficient routing strategy is crucial to the deployment of VANETs. This paper deals with the optimal parameter setting of the optimized link state routing (OLSR), which is a well-known mobile ad hoc network routing protocol, by defining an optimization problem. This way, a series of representative metaheuristic algorithms (particle swarm optimization, differential evolution, genetic algorithm, and simulated annealing) are studied in this paper to find automatically optimal configurations of this routing protocol. In addition, a set of realistic VANET scenarios (based in the city of Málaga) have been defined to accurately evaluate the performance of the network under our automatic OLSR. In the experiments, our tuned OLSR configurations result in better quality of service ( QoS ) than the standard request for comments (RFC 3626), as well as several human experts, making it amenable for utilization in VANET configurations.
\end{abstract}

Index Terms-Metaheuristics, optimization algorithms, optimized link state routing (OLSR), vehicular ad hoc networks (VANET).

\section*{I. Introduction}
VEHICULAR ad hoc networks (VANETs) [1] are selfconfiguring networks where the nodes are vehicles (equipped with onboard computers), elements of roadside infrastructure, sensors, and pedestrian personal devices. WiFi (IEEE 802.11-based) technologies are used for deploying such kind of networks. At present, the IEEE group is completing the IEEE 802.11p and IEEE 1609 final drafts, which are known as "Standard Wireless Access in Vehicular Environments" (WAVE), specifically designed for VANETs. This technology presents the opportunity to develop powerful car systems capable of gathering, processing, and distributing information.

For example, a driver assistance system could collect accurate and up-to-date data about the surrounding environment, detect potentially dangerous situations, and notify the driver [2].

In VANETs, the WiFi limitations in coverage and capacity of the channel, the high mobility of the nodes, and the presence of obstacles generate packet loss, frequent topology changes, and network fragmentation. Thus, a great deal of effort is dedicated to offer new medium access control access strategies [3] and to design efficient routing protocols [4], [5]. In turn, in such kind of networks, routing is a challenging task since there is no central entity in charge of finding the routing paths among the nodes. Different routing strategies have been defined based on prior ad hoc network architectures by targeting the specific VANET needs of scenarios and applications. These protocols can be grouped into topology based (proactive, e.g., destination-sequenced distance-vector and optimized link state routing (OLSR), reactive, e.g., ad hoc On demand distance vector (AODV) and dynamic source routing (DSR), carry-and-forwarding, etc.), position based (e.g., greedy perimeter stateless routing (GPSR) and greedy perimeter coordinator routing), cluster based (e.g., clustering for open IVC network and location-based routing algorithm with cluster-based flooding), and broadcasting (e.g., BROADCOMM and history enhanced vector based tracing detection) protocols [5].

Most of the VANET applications critically rely on routing protocols. Thus, an optimal routing strategy that makes better use of resources is crucial to deploy efficient VANETs that actually work in volatile networks. Finding well-suited parameter configurations of existing mobile ad hoc network (MANET) protocols is a way of improving their performance, even making the difference between a network that does work or does not, e.g., networks with high routing load suffer from congestion and cannot ensure timely and reliable delivery of messages [6].

In the present paper, we aim at defining and solving an offline optimization problem to efficiently and automatically tune OLSR [7], which is a widely used MANET unicast proactive routing protocol. Although specific routing protocols are emerging for VANET networks, a number of authors are currently using OLSR to deploy vehicular networks [8]-[11]. This protocol has been chosen mainly because it presents a series of features that make it well suited for VANETs. It exhibits very competitive delays in the transmission of data packets in large networks (which is an important feature for VANET applications), it adapts well to continuous topology changes, and the OLSR has simple operation that allows it to be easily integrated into different kinds of systems. More details about the use of OLSR in VANETs are provided in the following section.

However, the continuous exchange of routing control packets causes the appearance of network congestion, then limiting the global performance of the VANET. Thus, the quality-of-service (QoS) of OLSR significantly depends on the selection of its parameters, which determine the protocol operation. As shown in the results presented in [12] and in the present paper, OLSR has a wide range of improvements by changing the configuration parameters. Therefore, computing an optimal configuration for the parameters of this protocol is crucial before deploying any VANET, since it could decisively improve the QoS, with a high implication on enlarging the network data rates and reducing the network load. Then, all these features make OLSR a good candidate to be optimally tuned.

In this paper, we define an optimization problem to tune the OLSR protocol, obtaining automatically the configuration that best fits the specific characteristics of VANETs. An optimization problem is defined by a search space and a quality or fitness function. The search space restricts the possible configurations of a solution vector, which is associated with a numerical cost by the fitness function. Thus, solving an optimization problem consists of finding the least-cost configuration of a solution vector. In spite of the moderate number of configurable parameters that govern OLSR [7], the number of possible combinations of values that they can take makes this task very hard.

Due to the high complexity that these kinds of problems usually show, the use of automatic intelligent tools is a mandatory requirement when facing them. In this sense, metaheuristic algorithms [13] emerge as efficient stochastic techniques able to solve optimization problems. Indeed, these algorithms are currently employed in a multitude of engineering problems [14], [15], showing a successful performance. Unfortunately, the use of metaheuristics in the optimization of ad hoc networks (and concretely in VANETs) is still limited, and just a few related approaches can be found in the literature.

In [16], a specialized cellular multiobjective genetic algorithm was used for finding an optimal broadcasting strategy in urban MANETs. In [17], six versions of a genetic algorithm (GA) (panmictic and decentralized) were evaluated and used in the design of ad hoc injection networks. A GA was also employed by Cheng and Yang [18] for solving the multicast routing problem in MANETs. Due to its specific design, Shokrani and Jabbehdari [19] developed new routing protocols for MANETs based on ant colony optimization. Particle swarm optimization (PSO) algorithm has been used to manage the network resources by Huang et al. [20], proposing a new routing protocol based on this algorithm to make scheduling decisions for reducing the packet loss rate in a theoretical VANET scenario. More recently, García-Nieto et al. [21] optimized the file transfer service (vehicular data transfer protocol) in realistic VANET scenarios (urban and highway) by using different metaheuristic techniques.

We evaluate here four different techniques: 1) PSO [22]; 2) differential evolution (DE) [23]; 3) GA [24]; and 4) simulated annealing (SA) [25]. We have chosen these algorithms because they represent a wide and varied set of solvers for real parameter optimization based on different search/optimization schemes and strategies. The popular network simulator $n s-2$ [26] is used in the fitness function evaluation of the solutions\\
(tentative OLSR parameters) generated by the optimization algorithms to guide the search process. A set of VANET instances have been defined by using real data (road specification and mobility) concerning an urban area in the city of Málaga, Spain.

In summary, the main contributions of this paper are the following.

\begin{enumerate}
  \item We propose an optimization strategy in which a number of metaheuristic algorithms are (separately) coupled with a network simulator $(n s-2)$ to find quasi-optimal solutions.
  \item This optimization strategy is used in this paper to find as fine-tuned as possible configuration parameters of the OLSR protocol, although it could directly be used also for a number of other routing protocols (AODV, PROAODV, GPSR, FSR, DSR, etc.).
  \item We obtain OLSR configurations that automatically outperform the standard one and those used by human experts in the current state of the art.
  \item We generate a set of realistic VANET scenarios based in the real area of Málaga, Spain. These instances are publicly available online for the sake of future experiments.
\end{enumerate}

The remainder of this paper is organized as follows. In the next section, the OLSR routing protocol is introduced. Section III describes the optimization design followed to tackle the problem. Section IV presents the experiments carried out. Results, comparisons, and analyses are shown in Section V. Finally, conclusions and future work are considered in Section VI.

\section*{II. Problem Overview}
Exchanging up-to-date information among vehicles is the most salient feature of a VANET. To do so, the packets have to travel through the network from one node to the others, which is a complex task in networks having high mobility and no central authority. The routing protocol operates in the core of VANETs, finding updated paths among the nodes to allow the effective exchange of data packets. For this reason, this paper deals with the optimization of a routing protocol, specifically the OLSR protocol [7].

This protocol has been chosen since it presents a series of features that make it suitable for highly dynamic ad hoc networks and concretely for VANETs. These features are the following.

\begin{enumerate}
  \item OLSR is a routing protocol that follows a proactive strategy, which increases the suitability for ad hoc networks with nodes of high mobility generating frequent and rapid topological changes, like in VANETs [27], [28].
  \item Using OLSR, the status of the links is immediately known. Additionally, it is possible to extend the protocol information that is exchanged with some data of quality of the links to allow the hosts to know in advance the quality of the network routes.
  \item The simple operation of OLSR allows easy integration into existing operating systems and devices (including smartphones, embedded systems, etc.) without changing the format of the header of the IP messages [27].
  \item The OLSR protocol is well suited for high density networks, where most of the communication is concentrated between a large number of nodes (as in VANETs) [28].
  \item OLSR is particularly appropriate for networks with applications that require short transmission delays (as most of warning information VANET applications) [28].
  \item Thanks to its capability of managing multiple interface addresses of the same host, VANET nodes can use different network interfaces (WiFi, Bluetooth, etc.) and act as gateways to other possible network infrastructures and devices (as drivers and pedestrian smartphones, VANET base stations, etc.) [7].
\end{enumerate}

The main drawback of OLSR is the necessity of maintaining the routing table for all the possible routes. Such a drawback is negligible for scenarios with few nodes, but for large dense networks, the overhead of control messages could use additional bandwidth and provoke network congestion. This constrains the scalability of the studied protocol.

However, this precise performance of OLSR significantly depends on the selection of its parameters [9], [12], [29]. For example, the detection of topological changes can be adjusted by changing the time interval for broadcasting HELLO messages. Thus, computing an optimal configuration for the parameters of this protocol is crucial before deploying any VANET, since it could decisively improve the QoS, with a high implication on enlarging the network data rates and reducing the network load. In addition, we have not considered a target application in particular for this paper, although we are more interested on final end-user services like infotainment, vehicle-to-vehicle multiplayer gaming, content distribution and sharing, etc. Such services rely on peer-to-peer communications and therefore unicast routing protocols like OLSR.

All these features make OLSR a good candidate to be optimally tuned and justifies our election, but nothing prevents our methodology to be applied on new VANET protocols.

\section*{A. OLSR Protocol}
OLSR is a proactive link-state routing protocol designed for MANETs (VANETs), which show low bandwidth and high mobility. OLSR is a type of classical link-state routing protocol that relies on employing an efficient periodic flooding of control information using special nodes that act as multipoint relays (MPRs). The use of MPRs reduces the number of required transmissions [30].

OLSR daemons periodically exchange different messages to maintain the topology information of the entire network in the presence of mobility and failures. The core functionality is performed mainly by using three different types of messages: 1) HELLO; 2) topology control (TC); and 3) multiple interface declaration (MID) messages.

\begin{enumerate}
  \item HELLO messages are exchanged between neighbor nodes (one-hop distance). They are employed to accommodate link sensing, neighborhood detection, and MPR selection signaling. These messages are generated periodically, containing information about the neighbor nodes and about the links between their network interfaces.
\end{enumerate}

TABLE I\\
Main OLSR Parameters and RFC 3626 Specified Values

\begin{center}
\begin{tabular}{|l|l|l|}
\hline
Parameter & Standard Configuration & Range \\
\hline
HELLO\_INTERVAL & 2.0 s & $\mathbb{R} \in[1.0,30.0]$ \\
REFRESH\_INTERVAL & 2.0 s & $\mathbb{R} \in[1.0,30.0]$ \\
TC\_INTERVAL & 5.0 s & $\mathbb{R} \in[1.0,30.0]$ \\
WILLINGNESS & 3 & $\mathbb{Z} \in[0,7]$ \\
NEIGHB\_HOLD\_TIME & $3 \times$ HELLO\_INTERVAL & $\mathbb{R} \in[3.0,100.0]$ \\
TOP\_HOLD\_TIME & $3 \times$ TC\_INTERVAL & $\mathbb{R} \in[3.0,100.0]$ \\
MID\_HOLD\_TIME & $3 \times$ TC\_INTERVAL & $\mathbb{R} \in[3.0,100.0]$ \\
DUP\_HOLD\_TIME & 30.0 s & $\mathbb{R} \in[3.0,100.0]$ \\
\hline
\end{tabular}
\end{center}

\begin{enumerate}
  \setcounter{enumi}{1}
  \item TC messages are generated periodically by MPRs to indicate which other nodes have selected it as their MPR. This information is stored in the topology information base of each network node, which is used for routing table calculations. Such messages are forwarded to the other nodes through the entire network. Since TC messages are broadcast periodically, a sequence number is used to distinguish between recent and old ones.
  \item MID messages are sent by the nodes to report information about their network interfaces employed to participate in the network. Such information is needed since the nodes may have multiple interfaces with distinct addresses participating in the communications.\\[0pt]
The OLSR mechanisms are regulated by a set of parameters predefined in the OLSR RFC 3626 [7] (see Table I). These parameters have been tuned by different authors without using any automatic tool in [12] and [29], and they are the timeouts before resending HELLO, MID, and TC messages (HELLO\_INTERVAL, REFRESH\_INTERVAL, and TC\_INTERVAL, respectively); the "validity time" of the information received via these three message types, which are NEIGHB\_HOLD\_TIME (HELLO), MID\_HOLD\_TIME (MID), and TOP\_HOLD\_TIME (TC); the WILLINGNESS of a node to act as an MPR (to carry and forward traffic to other nodes); and DUP\_HOLD\_TIME, which represents the time during which the MPRs record information about the forwarded packets.
\end{enumerate}

\section*{B. OLSR Parameter Tuning}
The standard configuration of OLSR offers a moderate QoS when used in VANETs [11]. Hence, taking into account the impact of the parameter configuration in the whole network performance, we tackled here the problem of the optimal OLSR parameter tuning to discover the best protocol configuration previously to the deployment of VANET. The standard OLSR parameters are defined without clear values for their ranges. Table I shows the standard OLSR parameter values, as specified in the OLSR RFC 3626 [7]. The range of values each parameter can take has been defined here by following OLSR restrictions with the aim of avoiding pointless configurations.

According to that, we can use the OLSR parameters to define a solution vector of real variables, each one representing a given OLSR parameter. This way, the solution vector can automatically be fine-tuned by an optimization technique, with the aim of obtaining efficient OLSR parameter configurations for VANETs, hopefully outperforming the standard one defined in the RFC 3626 [7]. Additionally, analytic comparisons of\\
\includegraphics[max width=\textwidth, center]{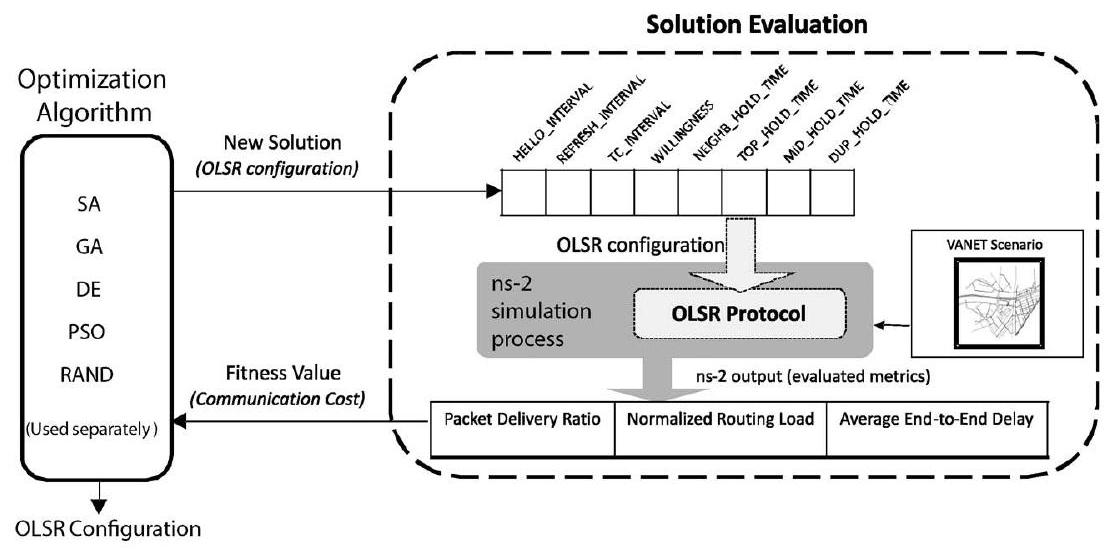}

Fig. 1. Optimization framework for automatic OLSR configuration in VANETs. The algorithms invoke the $n s-2$ simulator for solution evaluation.\\
different OLSR configurations and their performances as those done in this paper can help the experts identify the main source of communication problems and assist them in the design of new routing protocols.

To evaluate the quality or fitness of the different OLSR configurations (tentative solutions), we have defined a communication cost function in terms of three of the most commonly used QoS metrics in this area [9], [29]: 1) the packet delivery ratio (PDR), which is the fraction of data packets originated by an application that is completely and correctly delivered; 2) the network routing load (NRL), which is the ratio of administrative routing packet transmissions to data packets delivered, where each hop is counted separately; and finally, 3) the end-to-end delay (E2ED), which is the difference between the time a data packet is originated by an application and the time this packet is received at its destination.

\section*{III. Optimization Framework}
As previously commented, the optimization strategy used to obtain automatically efficient OLSR parameter configurations is carried out by coupling two different stages: 1) an optimization procedure and 2) a simulation stage. The optimization block is carried out by a metaheuristic method, in this case one of those previously mentioned, i.e., PSO, DE, GA, and SA. These four methods were conceived to find optimal (or quasioptimal) solutions in continuous search spaces, which is the case in this paper. We use a simulation procedure for assigning a quantitative quality value (fitness) to the OLSR performance of computed configurations in terms of communication cost. This procedure is carried out by means of the $n s-2$ [26] network simulator widely used to simulate VANETs accurately [31]. For this paper, $n s-2$ has been modified to interact automatically with the optimization procedure with the aim of accepting new routing parameters, opening the way for similar future research.

As Fig. 1 illustrates, when the used metaheuristic requires the evaluation of a solution, it invokes the simulation procedure of the tentative OLSR configuration over the defined VANET scenario. Then, $n s-2$ is started and evaluates the VANET under the circumstances defined by the OLSR routing parameters generated by the optimization algorithm. After the simulation,\\
$n s-2$ returns global information about the PDR, the NRL, and the E2ED of the whole mobile vehicular network scenario, where there were 10 independent data transfers among the vehicles. This information is used in turn to compute the communication cost (comm\_cost) function as follows:

\begin{equation*}
\text { comm_cost }=w_{2} \cdot N R L+w_{3} \cdot E 2 E D-w_{1} \cdot P D R \tag{1}
\end{equation*}

The communication cost function represents the fitness function of the optimization problem addressed in this paper. To improve the QoS, the objective here consists of maximizing the PDR and minimizing both NRL and E2ED. As expressed in (1), we used an aggregative minimizing function, and for this reason, PDR was formulated with a negative sign. In this equation, factors $w_{1}, w_{2}$, and $w_{3}$ were used to weigh the influence of each metric on the resultant fitness value. These values were set in a previous experimentation, although resulting in poor solutions with low PDR and high NRL. We observed that in VANETs (highly dynamic environments), the OLSR delivers a great number of administrative packets, which increases the NRL, hence damaging the PDR. Since we are interested in promoting the PDR for the sake of an efficient communication of packets, we decided in this approach to use different biased weighs in the fitness function, being $w_{1}=0.5, w_{2}=0.2$, and $w_{3}=0.3$. This way, PDR takes priority over NRL and E2ED since we first look for the routing effectiveness and second (but also important) for the communication efficiency.

\section*{IV. Experiments}
The simulation task should offer a network environment as close as possible to the real world environment. Following this idea, we make an effort to define realistic scenarios, where VANETs may be deployed. In this section, we define the urban scenario used in our simulations. Next, we present the experimental setup, taking into account the parameter settings for both the metaheuristic algorithms and the $n s-2$ simulation.

\section*{A. Urban VANET Scenario}
Since $n s-2$ is a network simulator of general purpose, it does not offer a way for directly defining realistic VANET simulations, where the nodes follow the behavior of vehicles\\
\includegraphics[max width=\textwidth, center]{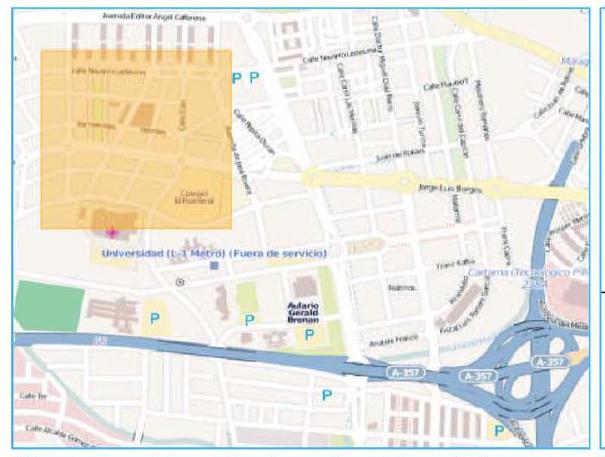}

Topology network of selected urban area\\
(SUMO view)\\
\includegraphics[max width=\textwidth, center]{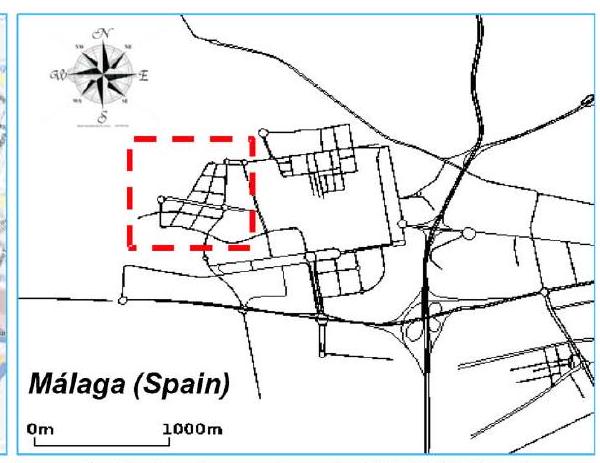}\\
*Real urban information is captured from digital map and exported to topology network in SUMO/NS2 instance\\
Fig. 2. Málaga real urban VANET scenario.\\[0pt]
in a road, traffic lights, traffic signs, etc. To solve this problem, we have used the Simulation of Urban MObility (SUMO) road traffic simulator to generate realistic mobility models [32]. This tool returns traces with the mobility definitions that can be used by $n s-2$. The main advantage of employing traffic simulators is that they can be used to generate realistic VANET environments by automatically selecting real areas from freely available digital maps (OpenStreetMap, ${ }^{1}$ as specified in Fig. 2), taking into account road directions, traffic lights and signs, etc. The VANET instance defined in this paper contains 30 cars moving through the roads selected of an area of $1200 \times$ $1200 \mathrm{~m}^{2}$ from the city downtown of Málaga (Spain) during 3 min. The area inside the dotted line box of Fig. 2 shows the roads taken into account to define the VANET urban scenario for our experiments. Through the simulation time, a set of cars exchange data, and as in an urban road, their speed fluctuates between $10 \mathrm{~km} / \mathrm{h}(2.78 \mathrm{~m} / \mathrm{s})$ and $50 \mathrm{~km} / \mathrm{h}(13.88 \mathrm{~m} / \mathrm{s})$.

For this VANET scenario, we have defined a specific data flow trustworthy representing different possible communications that may exist. The data flow model performs 10 sessions of a constant bit rate (CBR) data generator that operates over user datagram protocol (UDP) agents defined in the nodes (vehicles). This way, the interconnected vehicles exchange the data generated by the CBR agents. The CBR data packet size is 512 B , and the packet rate is 4 packets/s. The remaining simulation parameters are summarized in Table II for future reproduction purposes. We have chosen a fixed data rate since we do not aim to study the maximum throughput, but we want to investigate the ability of OLSR to successfully find and keep routes.

\section*{B. Experimental Setting}
The experiments have been carried out by using four metaheuristics (PSO, DE, GA, and SA). The implementation of these algorithms is provided by MALLBA [33]: a C++-based framework of metaheuristics for solving optimization problems. Additionally, we have employed a random search algorithm (RAND) also developed in $\mathrm{C}++$. To compare these

TABLE II\\
Main Parameters Defined for the $n s-2$ Simulation

\begin{center}
\begin{tabular}{|ll|}
\hline
Parameter & Value \\
\hline
Simulation time & 180 s \\
Simulation area & $1,200 \times 1,200 \mathrm{~m}^{2}$ \\
Number of vehicles & 30 \\
Vehicle speed & $10-50 \mathrm{~km} / \mathrm{h}$ \\
Propagation model & Two Ray Ground \\
Radio frequency & 2.47 GHz \\
Channel bandwidth & 5.5 Mbps \\
PHY/MAC protocols & IEEE 802.11 b \\
Max. Transmission Range & 250 ms \\
Max. N. of Retrans. MAC & 6 \\
Routing protocol & OLSR \\
Transport protocol & UDP \\
CBR data flow & 10 sessions \\
\hline
\end{tabular}
\end{center}

TABLE III\\
Parameterization of the Optimization Algorithms

\begin{center}
\begin{tabular}{|lllr|}
\hline
Algorithm & Parameter & Symbol & Value \\
\hline
\multirow{3}{*}{PSO} & Local Coefficient & $\varphi_{1}$ & 2 \\
 & Social Coefficient & $\varphi_{2}$ & 2 \\
 & Inertia Weigh & $w$ & 0.50 \\
\hline
\multirow{2}{*}{DE} & Crossover Probability & $C r$ & 0.90 \\
 & Mutation Factor & $\mu$ & 0.10 \\
\hline
\multirow{2}{*}{GA} & Crossover Probability & $P_{\text {cros }}$ & 0.80 \\
 & Mutation Probability & $P_{\text {mut }}$ & 0.01 \\
\hline
SA & Temperature Decay & $T$ & 0.80 \\
\hline
\end{tabular}
\end{center}

five methods in the OLSR parameter tuning, they have been executed to reach the same stop condition, i.e., 1000 fitness function evaluations. SA and RAND are performed 1000 iteration steps, and the population-based algorithms performed 100 generations with populations of 10 individuals $(100 \times 10=$ 1000) for each one of them. The main parameters of these algorithms are summarized in Table III. The simulation phase is carried out by running $n s-2$ simulator version $n s-2.34$ using the UM-OLSR (version 0.8.8) ${ }^{2}$ implementation of OLSR. We have performed 30 independent runs of every optimization technique on machines with Pentium IV $2.4-\mathrm{GHz}$ core, 1 GB of RAM, and O.S. Linux Fedora core 6.

TABLE IV\\
Results Obtained by Metaheuristic and Rands in the Optimization of OLSR for OUr VANET Scenario. Results of the Statistical Tests of Friedman and Kruskal-Wallis (KW) Are Also Provided

\begin{center}
\begin{tabular}{l|c|c|c|c||c|c}
\hline
Alg. & Mean $_{\text {std }}$ & Best & Median & Worst & Fried. & KW (p-value) \\
\hline
SA & $-\mathbf{0 . 4 5 0} \pm 0.024$ & -0.478 & $\mathbf{- 0 . 4 5 7}$ & $\mathbf{- 0 . 4 0 7}$ & $\mathbf{1 . 4 0}$ & $\mathbf{3 . 0 5 9 E - 6}$ \\
DE & $-0.437 \pm 0.030$ & -0.480 & -0.435 & -0.393 & 2.10 & $3.066 \mathrm{E}-6$ \\
PSO & $-0.432 \pm 0.033$ & $\mathbf{- 0 . 4 8 2}$ & -0.420 & -0.392 & 2.50 & $3.067 \mathrm{E}-6$ \\
GA & $-0.351 \pm \mathbf{0 . 0 2 3}$ & -0.437 & -0.345 & -0.327 & 4.33 & $3.159 \mathrm{E}-6$ \\
RAND & $-0.3308 \pm 0.050$ & -0.410 & -0.330 & -0.217 & 4.50 & $3.358 \mathrm{E}-6$ \\
\hline
\end{tabular}
\end{center}

\begin{center}
\includegraphics[max width=\textwidth]{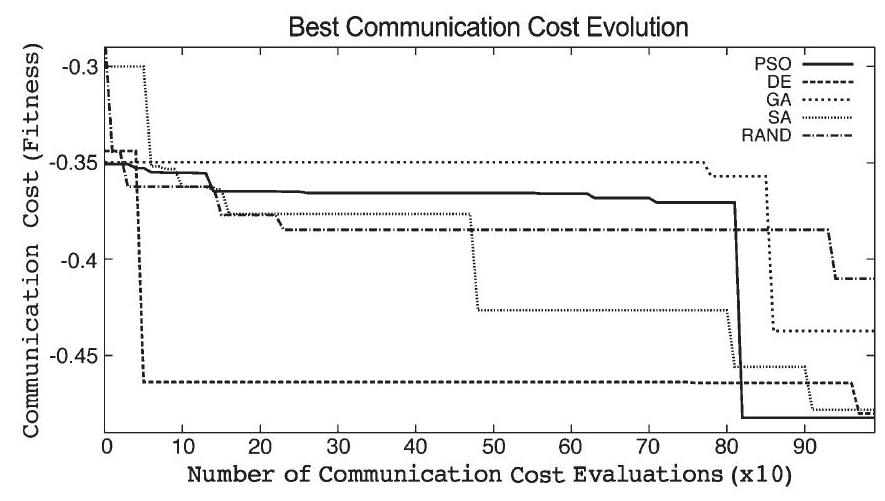}
\end{center}

Fig. 3. Metaheuristic and RAND best fitness value evolution when solving the OLSR configuration problem.

\section*{V. Results}
This section presents the experimental results from four different points of view. First, we show metaheuristic and RAND performances when solving the OLSR optimization problem. Second, we compare the obtained OLSR parameter configurations against several configurations found in the literature. Third, the obtained solutions are also evaluated on multiple different scenarios to check whether optimized OLSR parameters can be used in general or not. Fourth, we analyze the influence of the different OLSR parameters in the global QoS provided by the network.

\section*{A. Performance Analysis}
Table IV shows the mean and standard deviation of the communication cost values obtained (out of 30 independent executions) running all the evaluated metaheuristics and the RAND for the VANET scenario instance. The best, median, and worst values are also provided.

We can clearly observe in this table that SA outperformed all the other algorithms in terms of mean ( -0.450297 ), median ( -0.457451 ), and worst ( -0.406932 ) communication cost values. According to these measures, SA obtained the best results, followed by DE, PSO, and GA, respectively. Finally, as expected, the RAND obtains worse results than all the metaheuristic algorithms.

In terms of the best OLSR configuration returned by the algorithms (third column), PSO computed the solution with the lowest communication cost (see Fig. 3). The best OLSR parameter settings obtained by $\mathrm{DE}, \mathrm{SA}$, and GA are the second, third, and forth, respectively. The RAND best OLSR configuration is the least competitive one.

TABLE V\\
Mean Execution Time per Independent Run of Each Algorithm

\begin{center}
\begin{tabular}{l|cc}
\hline
Algorithm & $T_{\text {best }}$ (seconds) & $T_{\text {run }}$ (seconds) \\
\hline
PSO & $3.05 \mathrm{E}+04$ & $5.38 \mathrm{E}+04$ \\
DE & $4.29 \mathrm{E}+04$ & $7.95 \mathrm{E}+04$ \\
GA & $\mathbf{2 . 0 4 E}+\mathbf{0 4}$ & $1.18 \mathrm{E}+05$ \\
SA & $5.78 \mathrm{E}+04$ & $1.04 \mathrm{E}+05$ \\
RAND & $3.73 \mathrm{E}+04$ & $\mathbf{4 . 3 6 E}+\mathbf{0 4}$ \\
\hline
\end{tabular}
\end{center}

TABLE VI\\
Rankings in Terms of Best Returned Solution, Time to Find the Optimum $T_{\text {best }}$, and Mean Execution Time $T_{\text {run }}$

\begin{center}
\begin{tabular}{l|r|rr}
\hline
Rank & Mean $_{\text {fitness }}$ & $T_{\text {best }}$ & $T_{\text {run }}$ \\
\hline
1 & SA & GA & RAND \\
2 & DE & PSO & PSO \\
3 & PSO & RAND & DE \\
4 & GA & DE & SA \\
5 & RAND & SA & GA \\
\hline
\end{tabular}
\end{center}

With the aim of providing these comparisons with statistical confidence, we have applied the Friedman and Kruskal-Wallis tests [34] to the distributions of the results. We have used these nonparametric tests since the resulting distributions violated the conditions of equality of variances several times. The confidence level was set to $95 \%(p-$ value $=0.05)$, which allows us to ensure that all these distributions are statistically different if they result in $p-$ value $<0.05$.

In effect, confirming the previous observations, the results of Friedman (see sixth column of Table IV) test ranked SA as the algorithm with the best global performance followed by DE, PSO, and GA, respectively. The random search (RAND) showed the worst rank among the compared techniques. Moreover, the multicompare test of Kruskal-Wallis applied to the median values of the distributions resulted in $p$-values $\ll$ 0.05 (last column of Table IV). Therefore, we can claim that all the compared algorithms obtained statistically different results.

According to the behavior of the optimization algorithms, we now study the evolution of the best solution (communication cost value) during the whole evolutionary process. Fig. 3 plots the graph of the best communication cost (fitness value) tracked throughout the best execution for each algorithm. We can observe that DE, PSO, and SA converge in the same range of solutions. However, their evolution is different. The major improvement of the DE solution occurs during the first steps of the execution, unlike what happens with PSO that improves its solution during the last steps. SA, the best ranked algorithm according to the Friedman test, performs several gradual solution improvements during the whole execution.

Finally, concerning the mean run time that each algorithm spent in the experiments, Table V shows the mean time in which the best solution was found $T_{\text {best }}$ (second column) and the global mean run time $T_{\text {run }}$ (third column).

GA shows the shortest time $(2.04 \mathrm{E}+04 \mathrm{~s})$ to find its best solutions, and it seems that this algorithm quickly falls in local optima, hence obtaining weak results (see Table IV). Globally, PSO needed the second shortest time $(3.05 \mathrm{E}+04 \mathrm{~s})$ to compute its optima followed by DE, RAND, and SA, respectively.

In terms of mean run time $T_{\text {run }}$, random search takes shorter times $(4.36 \mathrm{E}+04 \mathrm{~s})$ than the other algorithms since it has less internal operations. However, this algorithm converges to

TABLE VII\\
OLSR Parameter Values in Configurations of the State of the Art (Gomez etal.), the Standard RFC 3626, and the Best Solutions in Optimization Algorithms

\begin{center}
\begin{tabular}{|c|c|c|c|c|c|c|c|c|c|}
\hline
 & \multicolumn{3}{|l|}{Gómez et al. [12]} & \multirow[t]{2}{*}{\(
\begin{array}{r}
\hline \text { OLSR } \\
\text { RFC }
\end{array}
\)} & \multirow[t]{2}{*}{RAND} & \multicolumn{4}{|l|}{Optimized Configurations} \\
\hline
Parameter & \#1 & \#2 & \#3 &  &  & DE & PSO & GA & SA \\
\hline
HEL\_INT & 0.50 & 1.0 & 4.0 & 2.0 & 3.730 & 8.477 & 8.909 & 8.568 & 9.005 \\
\hline
REFR\_INT & 0.50 & 1.0 & 4.0 & 2.0 & 6.188 & 1.086 & 9.663 & 15.829 & 4.925 \\
\hline
TC\_INT & 1.25 & 2.5 & 10.0 & 5.0 & 5.188 & 7.246 & 7.192 & 5.286 & 6.753 \\
\hline
WILLING. & 3 & 3 & 3 & 3 & 4 & 0 & 1 &  & 0 \\
\hline
NEIG\_H\_T & 1.50 & 3.0 & 12.0 & 6.0 & 5.400 & 16.924 & 67.238 & 83.771 & 80.334 \\
\hline
TOP\_H\_T & 3.75 & 7.5 & 20.0 & 15.0 & 40.164 & 99.061 & 72.693 & 67.619 & 80.965 \\
\hline
MID\_H\_T & 3.75 & 7.5 & 20.0 & 15.0 & 34.476 & 6.713 & 91.303 & 37.105 & 2.913 \\
\hline
DUP\_H\_T & 30.0 & 30.0 & 30.0 & 30.0 & 31.515 & 71.938 & 21.572 & 16.268 & 16.705 \\
\hline
\end{tabular}
\end{center}

TABLE VIII\\
QoS Comparisons of Considered OLSR Configurations

\begin{center}
\begin{tabular}{|c|c|c|c|c|c|c|c|c|c|}
\hline
 & \multicolumn{3}{|l|}{Gómez et al. [12]} & \multirow[t]{2}{*}{\(
\begin{array}{r}
\text { OLSR } \\
\text { RFC }
\end{array}
\)} & \multirow[t]{2}{*}{RAND} & \multicolumn{4}{|l|}{Optimized Configurations} \\
\hline
Metric & \#1 & \#2 & \#3 &  &  & DE & PSO & GA & SA \\
\hline
PDR (\%) & 71.43 & 87.50 & 93.34 & 91.67 & 94.12 & 100.00 & 100.00 & 100.00 & 100.00 \\
\hline
NRL (\%) & 89.54 & 32.48 & 14.13 & 9.52 & 6.93 & 2.71 & 2.90 & 13.12 & 4.84 \\
\hline
E2ED(ms) & 5.41 & 5.03 & 7.19 & 6.29 & 6.57 & 15.60 & 11.31 & 19.17 & 4.73 \\
\hline
\end{tabular}
\end{center}

the weakest solutions. PSO is the metaheuristic that spent the shortest mean running time $(5.38 \mathrm{E}+04 \mathrm{~s})$ followed by DE, SA, and GA, respectively.

According to Table V, the analyzed algorithms use between $4.36 \mathrm{E}+04$ and $1.18 \mathrm{E}+05$ seconds $(12.11$ and 32.66 h , respectively) to finish each execution. This effort in the protocol design is completely justified by the subsequent benefits obtained in the global quality of service once the VANET is physically deployed (see Section V-B).

Table VI summarizes three rankings of the compared algorithms in terms of the mean fitness quality $M e a n_{\text {fitness }}$ when solving the OLSR optimization problem, the mean time in which the best solution was found $T_{\text {best }}$, and the mean run time $T_{\text {run }}$. In light of these results, we can claim that SA performs the best in terms of quality of solutions, although spending a higher time than other algorithms like PSO and DE to converge to such accurate solutions. In contrast, PSO offers the best tradeoff between the solution quality and the time required to find it. This is in fact a typical behavior of PSO [35], where the early convergence to successful solutions it performs makes this algorithm useful for time consuming problems as the choice for the present paper. This way, we can offer accurate OLSR configurations to experts in reasonable design times.

Thus, the best algorithm if there is not any time restriction is SA, since it is the one ranked with the best global performance. However, PSO offers the best tradeoff between the time requirements and the quality of the returned solution.

\section*{B. Optimized Versus Human Expert Configurations}
In this section, we focus our analysis to the solution domain point of view. Then, we compare the resulted OLSR configurations in terms of selected QoS indicators ( $P D R, N R L$, and $E 2 E D)$. To start with, we can see in Table VII the OLSR parameter settings considered for comparison in this analysis. In this table, columns 2-4 contain three human expert configurations (\#1, \#2, and \#3), as proposed by Gómez et al. [12]; columns 5 and 6 contain the OLSR configurations of the standard RFC 3626 and the one obtained by the random search, respectively;\\
and columns $7-10$ show the best OLSR configurations obtained by each of the four metaheuristic algorithms studied in this paper: PSO, DE, GA, and SA.

The results of simulating our VANET scenario instance of Málaga city with these OLSR configurations are presented in Table VIII. Our observations are the following.

\begin{enumerate}
  \item Examining the PDR indicator, we can effectively check that the four metaheuristic algorithms obtained a $100 \%$ contrast with the RAND that achieved $94.12 \%$ and the remaining configurations that obtained PDRs between $71.43 \%$ and $93.34 \%$. This is an important issue in highly dynamic VANETs since a low PDR directly implies a higher packet loss, which makes the OLSR protocol generate additional administrative packets with an impact in the network congestion.
  \item Concerning the NRL, similar results can be observed. That is, almost all the optimized OLSR configurations showed better routing loads than the other proposals. Only GA (the worst-ranked metaheuristic) obtained an NRL (13.12\%) worse than the two obtained by the standard configuration and RAND ( $9.52 \%$ and $6.93 \%$, respectively), although it was better than the three human expert configurations (\#1 with $89.54 \%$, \#2 with $32.48 \%$, and \#3 with $14.13 \%$ ). In general, DE generated the lowest routing load ( $2.71 \%$ ) followed by PSO ( $2.90 \%$ ). These results contrast with all the other configurations since DE and PSO outperformed the rest by one order of magnitude (two in the case of \#1). Reducing the routing load is important since this is a way to reduce the possibility of network failures related to the congestion problem in VANETs [6].
  \item Finally, in terms of the E2ED, we can notice that SA obtained the best result ( 4.73 ms ), followed by human expert configurations (Gómez et al.), standard RFC 3626, and RAND. In this case, the remaining metaheuristic algorithms (PSO, DE, and GA) showed a moderate performance. Evidently, the low routing load experimented in these configurations limited the routing management\\
\includegraphics[max width=\textwidth, center]{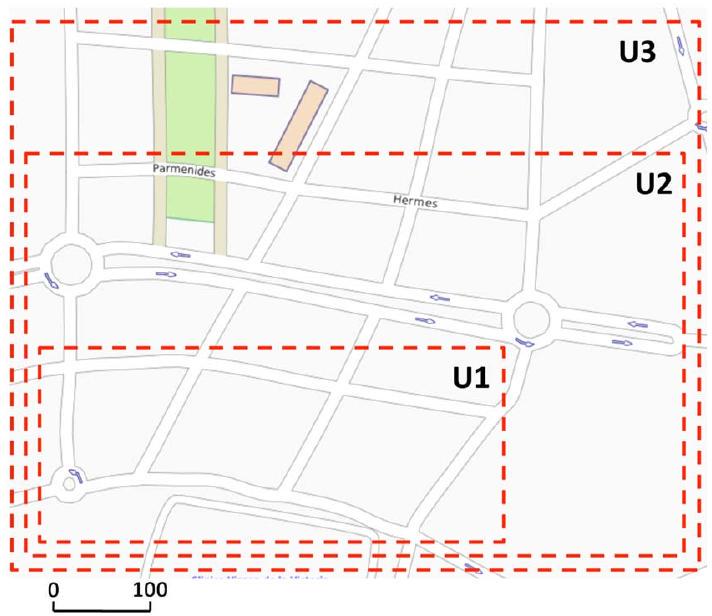}
\end{enumerate}

Fig. 4. Three different scaling areas of the selected network topology.\\[0pt]
operations, hence making the average E2ED worse than the other configurations with high routing load. However, it is remarkable that all the optimized OLSR parameter settings analyzed here send the packets with a delay shorter than 20 ms , which is the highest allowed latency for cooperative vehicular applications [2], the most critical ones.

\section*{C. Optimized Configurations on Multiple Scenarios}
Once we have shown the benefits of using our approach on the optimized scenario, in this section, we evaluate the obtained OLSR configurations on multiple different scenarios. This way, we aim to validate optimized parameters on different conditions of traffic density, network use, and area dimension. For this task, we have carried out an extensive set of validation experiments including the simulation of 54 different urban VANET scenarios. Then, we evaluate the results in terms of four routing QoS metrics: $P D R, N R L, E 2 E D$, and routing path length (RPL).

For the scenario definitions, three different geographical area sizes have been selected from the downtown of Málaga by using SUMO to study the scalability of our approach (see Fig. 4). Additionally, we have extended our analysis by studying how do various road traffic situations affect the routing performance. Thus, we have generated three different road traffic densities (number of vehicles moving through the roads) for each geographical area: $L$ (low), $M$ (medium), and $H$ (high). In these scenarios, $L$ represents traffic situations with the lowest number of vehicles so that the vehicles have greater freedom of mobility. On the contrary, $H$ has the largest number of vehicles, and it suffers from some situations of road traffic congestions, mainly at the crossroads. Finally, $M$ is an intermediate traffic situation between $L$ and $H$. The vehicles move through the roads during 180 s .

In each scenario, a number of unicast data transfers are carried out by pairs of VANET nodes that use one-hop and multihop communications. The vehicles communicate with each other by exchanging the data generated by a CBR generator during 30 s . The number of vehicles that generates the information (CBR sources) to be sent to the other nodes\\
depends on the VANET scenario (see Table IX). In turn, we have defined different VANET instances by using six different network data loads (traffic data rates) to analyze the capacity of our routing approaches with different workloads. Those are grouped in low rates ( 33,66 , and $100 \mathrm{~kb} / \mathrm{s}$ ) and high rates ( 333 , 666 , and $1000 \mathrm{~kb} / \mathrm{s}$ ). The vehicles network devices employ one of the evaluated OLSR parameterizations to compute the routing paths among the VANET nodes. The VANET nodes are set with UDP as the transport layer protocol. The physical and data link layers are tuned by following the specifications of the IEEE 802.11p standard by using Mac802\_11Ext and WirelessPhyExt $n s-2$ modules. As in [36], we have included in the simulations the fading Nakagami radio propagation model representing the WAVE radio propagation in urban scenarios [37]. Table IX summarizes the main features of the network used in our VANET simulations. All these VANET instances are publicly available online for the sake of future experiments (\href{http://neo.lcc.uma.es/vanet}{http://neo.lcc.uma.es/vanet}).

Table X presents for each OLSR configuration found using metaheuristics and RANDs the median values for each studied metric, computed in the simulations performed over the 54 different VANET scenarios. The results are compared with the values obtained in the simulations performed with the standard OLSR configuration suggested by RFC 3626 . The best median values obtained for each metric are marked in bold. From these results, a series of observations are made as follows.

\begin{enumerate}
  \item Concerning the PDR metric, we check that the number of packets delivered is generally reduced with the size of the geographical area. The GA and SA configurations obtained the best PDR in the U1 scenario (99.95\%), PSO in the U2 scenario ( $86 \%$ ), and RFC in the U3 scenario ( $86.71 \%$ ). Globally, in terms of median results, the highest PDR is obtained by the RFC configuration (89.56\%). However, the differences between the performances of all the configurations in terms of this metric are just between $1 \%$ and $5 \%$.
  \item In terms of routing workload (NRL), we can observe from Table X that the configurations obtained by SA and DE show better values, so they can even decrease the NRL along with the scenario size. In particular, scenario U2 seems to be a source of high routing loads since, practically, for all solutions (except those of SA and DE), this indicator is increased. In general, optimized OLSR configurations improve the NRL since for the three scenarios, the standard configuration of RFC shows the worst value (overall NRL $=23.15 \%$ ), and it is twice that obtained by DE , which is the best one ( $11.98 \%$ ).
  \item If we examine the E2ED, we observe that the required time to deliver the packets is higher with the scenario area dimension. The OLSR configuration optimized by GA required the shortest E2ED in U1 and U2 scenarios with 2.10 and 3.81 ms , respectively. In scenario U3, DE obtained the best E2ED ( 19.19 ms ). Globally, the shortest median E2ED is obtained by SA with 4.04 ms .
  \item In terms of the computed routing paths (RPL), GA obtained the shortest paths in the U1 scenario, and the RFC configuration used the shortest paths in scenarios U2 and
\end{enumerate}

TABLE IX\\
Details of the Vanet Scenarios and Network Specification

\begin{center}
\begin{tabular}{|c|c|c|c|}
\hline
Scen. & Area size & Vehicles & \( \begin{gathered} \text { CBR } \\ \text { sources } \end{gathered} \) \\
\hline
U1 & 120,000 m ${ }^{2}$ & \( \begin{aligned} & 1 \frac{10}{}-1 \frac{5}{2} \\ & --{ }_{2} \\ \hline & \end{aligned} \) & \( \begin{array}{r} \hline \\ --\quad-8 \\ --\frac{8}{10} \\ \hline \end{array} \) \\
\hline
U2 & 240,000 m ${ }^{2}$ & \( \begin{aligned} & --\frac{20}{3 \overline{0}} \\ & --4 \overline{0} \end{aligned} \) & \( \begin{array}{r} 10 \\ --\frac{10}{15} \\ --\frac{20}{20} \\ \hline \end{array} \) \\
\hline
U3 & 360,000 $\mathrm{m}^{2}$ & \( ---\frac{30}{45} \) & \( \begin{array}{r} -\quad \frac{15}{23} \\ --\frac{3}{30} \\ \hline \end{array} \) \\
\hline
\end{tabular}
\end{center}

TABLE X\\
Median Results of the Validation Experiments

\begin{center}
\begin{tabular}{|c|c|c|c|c|c|}
\hline
Scenario & Configurations & $P D R$ & NRL & $E 2 E D$ & $R P L$ \\
\hline
\multirow{6}{*}{U1} & SA & 99.95\% & 15.09\% & 2.13 ms & 1.03 \\
\hline
 & DE & 92.58\% & 12.64\% & 4.34 ms & 1.09 \\
\hline
 & GA & 99.95\% & 16.95\% & 2.10 ms & 1.01 \\
\hline
 & PSO & 99.39\% & 12.73\% & 2.60 ms & 1.05 \\
\hline
 & $\overline{\mathrm{R}} \mathrm{AN} \overline{\mathrm{D}}$ & $9 \overline{4.71 \%}$ & 18.35\% & $1 \overline{7} .1 \overline{6} \mathrm{~ms}$ & $1.3 \overline{8}$ \\
\hline
 & RFC & 99.40\% & 22.28\% & 2.79 ms & 1.05 \\
\hline
\multirow{6}{*}{U2} & SA & 84.01\% & 12.36\% & 8.99 ms & 1.59 \\
\hline
 & DE & 85.77\% & 10.04\% & 9.23 ms & 1.54 \\
\hline
 & GA & 85.70\% & 15.82\% & 3.81 ms & 1.63 \\
\hline
 & PSO & 86.03\% & 12.36\% & 10.56 ms & 1.54 \\
\hline
 & $\overline{\mathrm{R}} \mathrm{AN} \overline{\mathrm{D}}$ & $84.98 \%$ & $\overline{2} 0.15 \%$ & $2 \overline{1} .1 \overline{1} \mathrm{~ms}$ & $1.4 \overline{1}$ \\
\hline
 & RFC & 85.91\% & 23.94\% & 8.27 ms & 1.18 \\
\hline
\multirow{6}{*}{U3} & SA & 74.85\% & 10.05\% & 44.38 ms & 1.29 \\
\hline
 & DE & 78.29\% & 9.07\% & 19.19 ms & 1.31 \\
\hline
 & GA & 75.08\% & 11.17\% & 44.95 ms & 1.30 \\
\hline
 & PSO & 75.05\% & 9.97\% & 45.81 ms & 1.28 \\
\hline
 & $\overline{\mathrm{R}} \mathrm{A} \overline{\mathrm{N}} \overline{\mathrm{D}}^{-}$ & 69.76\% & 15.5 8 \% & $\overline{3} 9 \overline{8} .4 \overline{2} \overline{\mathrm{~ms}}$ & $1.4 \overline{9}$ \\
\hline
 & RFC & 86.71\% & 20.65\% & 70.26 ms & 1.05 \\
\hline
\multirow{6}{*}{Global Results} & SA & 84.76\% & 14.56\% & 4.04 ms & 1.35 \\
\hline
 & DE & 84.29\% & 11.98\% & 10.24 ms & 1.34 \\
\hline
 & GA & 87.85\% & 16.32\% & 4.36 ms & 1.34 \\
\hline
 & PSO & 86.73\% & 12.73\% & 8.12 ms & 1.46 \\
\hline
 & $\overline{\mathrm{R}}$ AN $\overline{\mathrm{D}}$ & $8 \overline{88} . \overline{93} \%$ & 19.21\% & $1 \overline{7} .1 \overline{6} \mathrm{~ms}$ & $1.3 \overline{8}$ \\
\hline
 & RFC & 89.56\% & 23.15\% & 6.06 ms & 1.09 \\
\hline
\end{tabular}
\end{center}

U3. In general, the median path lengths obtained by the standard parameters of OLSR required the lower number of hops. In this case, the higher frequency of routing information exchange maintains the routing tables up to date, although generating a higher routing load.\\
In summary, we can confirm that automatically tuned OLSR configurations by metaheuristics offer the best tradeoff between the four QoS metrics in the scope of the multiple scenario conditions analyzed here. The solutions obtained by metaheuristic algorithms show high rates of packet delivery ( $>84 \%$ ) and low values of routing load $(<16.5 \%)$, end to end delays ( $<10.3 \mathrm{~ms}$ ), and routing paths lengths ( $<2 \mathrm{hops}$ ). The standard RFC solution also reached accurate median values of PDF ( $88.9 \%$ ) but with the drawback of a high routing load ( $>23 \%$ ), which is a critical concern in this kind of network.

\section*{D. Global QoS Analysis}
In this section, we analyze the impact of some interesting OLSR parameters in the global performance of a given configuration. Table VII presents the configurations taken into account in this paper.

\begin{center}
\begin{tabular}{|c|c|}
\hline
Parameter & Value/Protocol \\
\hline
Propagation model & Nakagami \\
\hline
Carrier Frequency & 5.89 GHz \\
\hline
Channel bandwidth & 6 Mbps \\
\hline
PHY/MAC Layer & IEEE 802.11p \\
\hline
Routing Lāēer & $\overline{\mathrm{OLSR}}$ \\
\hline
Transport Layer & $\bar{U} \mathrm{D} \overline{\mathrm{P}}$ \\
\hline
$\overline{\mathrm{CBR}}$ - Pācket $\overline{\text { Size }}$ & $\overline{512}$ bytes \\
\hline
CBR Data Rate & $33 / 66 / 100 \mathrm{kbps}$ \\
\hline
CBR Data Rate & 333/666/1000 kbps \\
\hline
CBR Time & 60 s \\
\hline
\end{tabular}
\end{center}

Generally, according to our tests, reducing the HELLO\_INTERVAL and TC\_INTERVAL parameters may help to improve the OLSR reactivity to route changes and link failures; for this reason, the shorter the HELLO and TC intervals, the lower the E2ED (see Gómez et al. [12], as well as RFC and RAND OLSR parameter settings in Tables VII and VIII). Nevertheless, this causes an increase in protocol load (NRL) being very likely to cause network congestion and packet loss (lower PDR), that is, reducing the number of nodes that can be accurately communicating in the network. In fact, as TC messages are broadcast to the whole network, the NRL is reduced critically by increasing TC\_INTERVAL parameter (see DE and PSO OLSR parameter settings in Table X).

Something similar happens with the NEIGHB\_ HOLD\_TIME and the TOP\_HOLD\_TIME timers because in the optimized OLSR they are larger than the other ones. The need for protocol management information update seems to be lower than that proposed by Gómez and in OLSR 3626 RFC configurations, since the communications carried out by using the PSO, DE, GA, and SA OLSR settings correctly exchanged more than $84 \%$ of the data packets within a time of smaller than 20 ms (highest allowed latency).

\section*{VI. Conclusion}
In this paper, we have addressed the optimal parameter tuning of the OLSR routing protocol to be used in VANETs by using an automatic optimization tool. For this task, we have defined an optimization strategy based on coupling optimization algorithms (PSO, DE, GA, and SA) and the $n s-2$ network simulator. In addition, we have compared the optimized OLSR configurations with the standard one in RFC 3626 as well as with human expert configurations found in the current state of the art. In turn, we have validated the optimized configurations found by comparing them with each other and with the standard tuning in RFC 3626 and studying their performance in terms of QoS over 54 VANET scenarios. These urban VANET instances are based on real data of the downtown of Málaga. In the light of the experimental results, we can conclude the following.

\begin{enumerate}
  \item In terms of the performance of the optimization techniques used in this paper, SA outperforms the other studied metaheuristic algorithms when solving the defined OLSR optimization problem because it is the best ranked after the Friedman test. However, PSO presents the best tradeoff between the performance and the execution time\\
requirements. In turn, a parallel version of PSO running in multiple processors can also further reduce the computational time derived from large VANET simulations. This way, we can offer accurate OLSR configurations to experts in reasonable design times.
  \item When using the automatically tuned configurations over the VANET scenario employed during the optimization task, all the packets are delivered correctly $(\mathrm{PDR}=$ $100 \%$ ), increasing the PDR regarding the standard configuration by $8.34 \%$ and between $6.66 \%$ and $28.57 \%$ regarding the other expert-defined configurations. In turn, the use of the optimized configurations dramatically reduces the routing load generated by OLSR.
  \item Globally, the validation experiments show that the optimized configurations reduced the network workload, generating about the half of the routing load than the RFC 3626 configuration. By reducing the routing load, the routing tables are updated less frequently, calculating routing paths $27 \%$ longer than the standard version. Nevertheless, the mitigation of the OLSR-related congestion problems by optimized configurations generally allowed to shorten the packet delivery times. In turn, these features were obtained while keeping the degradation of amount of delivered data lower than $5 \%$.
  \item According to these results, the automatically tuned OLSRs by using metaheuristics are more scalable than the standard version because they are less likely to be affected by medium access and congestion problems. Specifically, the PSO obtained configuration obtained the best tradeoff between the QoS and the routing workload.\\
The optimization methodology presented in this paper (coupling metaheuristics and a simulator) offers the possibility of automatically and efficiently customizing any protocol for any VANET scenario. We can now provide the experts in this area with an optimization tool for the configuration of communication protocols in the scope of VANETs. Currently, several research projects, i.e., European (CARLINK \href{http://carlink.lcc.uma.es}{http://carlink.lcc.uma.es}) National (RoadMe \href{http://roadme}{http://roadme}. \href{http://lcc.uma.es}{lcc.uma.es}), and Regional (DIRICOM \href{http://diricom.lcc}{http://diricom.lcc}. \href{http://uma.es}{uma.es}), are taking the most of our approach.
\end{enumerate}

As a matter of future work, we are currently extending our experiments with new still larger urban and highway VANET instances. In this line, we are tackling the problem with parallel versions of metaheuristic algorithms to solve time-consuming issues derived from large simulations. We are also defining new optimized configuration schemes for other communication protocols such as WAVE, UDP, etc., which should efficiently support actual VANET design. Concerning the improvement of the QoS of the network, we are considering the use of infrastructure nodes, the Global Positioning System, security protocols, and sensing information. Finally, we are planning new real tests (using vehicles traveling through different kinds of roads) to validate our simulations.

\section*{REFERENCES}
[1] H. Hartenstein and K. Laberteaux, VANET Vehicular Applications and Inter-Networking Technologies. Upper Saddle River, NJ: Wiley, Dec. 2009, ser. Intelligent Transport Systems.\\[0pt]
[2] The Crash Avoidance Metrics Partnership (CAMP) Vehicle Safety Communications Consortium, "Vehicle Safety Communications Consortium (2005). Vehicle safety communications project task 3 final report: Identify intelligent vehicle applications enabled by Dedicated Short Range Communications (DSRC)." Tech. Rep. 809859, National Highway Traffic Safety Administration, Office of Research and Development (USDOT), Washington, DC, 2005.\\[0pt]
[3] N. Lu, Y. Ji, F. Liu, and X. Wang, "A dedicated multi-channel MAC protocol design for VANET with adaptive broadcasting," in Proc. IEEE WCNC, Apr. 2010, pp. 1-6.\\[0pt]
[4] T. Taleb, E. Sakhaee, A. Jamalipour, K. Hashimoto, N. Kato, and Y. Nemoto, "A stable routing protocol to support ITS services in VANET networks," IEEE Trans. Veh. Technol., vol. 56, no. 6, pp. 3337-3347, Nov. 2007.\\[0pt]
[5] F. Li and Y. Wang, "Routing in vehicular ad hoc networks: A survey," IEEE Veh. Technol. Mag., vol. 2, no. 2, pp. 12-22, Jun. 2007. [Online]. Available: \href{http://dx.doi.org/10.1109/MVT.2007.912927}{http://dx.doi.org/10.1109/MVT.2007.912927}\\[0pt]
[6] W. Zhang, A. Festag, R. Baldessari, and L. Le, "Congestion control for safety messages in VANETs: Concepts and framework," in Proc. 8th ITST, Oct. 2008, pp. 199-203.\\[0pt]
[7] T. Clausen and P. Jacquet, "Optimized link state routing protocol (OLSR)," IETF RFC 3626, 2003. [Online]. Available: \href{http://www}{http://www}. \href{http://ietf.org/rfc/rfc3626.txt}{ietf.org/rfc/rfc3626.txt}\\[0pt]
[8] T. Chen, O. Mehani, and R. Boreli, "Trusted routing for VANET," in Proc. 9th Int. Conf. ITST, M. Berbineau, M. Itami, and G. Wen, Eds., Oct. 2009, pp. 647-652.\\[0pt]
[9] J. Haerri, F. Filali, and C. Bonnet, "Performance comparison of AODV and OLSR in VANETs urban environments under realistic mobility patterns," in Proc. 5th Annu. Med-Hoc-Net, S. Basagni, A. Capone, L. Fratta, and G. Morabito, Eds., Lipari, Italy, Jun. 2006.\\[0pt]
[10] A. Laouiti, P. Mühlethaler, F. Sayah, and Y. Toor, "Quantitative evaluation of the cost of routing protocol OLSR in a Vehicle ad hoc NETwork (VANET)," in Proc. VTC, 2008, pp. 2986-2990.\\[0pt]
[11] J. Santa, M. Tsukada, T. Ernst, O. Mehani, and A. F. Gómez-Skarmeta, "Assessment of VANET multi-hop routing over an experimental platform," Int. J. Internet Protocol Technol., vol. 4, no. 3, pp. 158-172, Sep. 2009.\\[0pt]
[12] C. Gómez, D. García, and J. Paradells, "Improving performance of a real ad hoc network by tuning OLSR parameters," in Proc. 10th IEEE ISCC, 2005, pp. 16-21.\\[0pt]
[13] C. Blum and A. Roli, "Metaheuristics in combinatorial optimization: Overview and conceptual comparison," ACM Comput. Surv., vol. 35, no. 3, pp. 268-308, Sep. 2003.\\[0pt]
[14] K. Parsopoulos and F. Vrahatis, "Unified particle swarm optimization for solving constrained engineering optimization problems," in Advances in Natural Computation. New York: Springer-Verlag, 2005, pp. 582-591.\\[0pt]
[15] E. Alba, J. García-Nieto, J. Taheri, and A. Zomaya, "New research in nature inspired algorithms for mobility management in GSM networks," in Proc. EvoWorkshops-LNCS, 2008, pp. 1-10.\\[0pt]
[16] E. Alba, B. Dorronsoro, F. Luna, A. Nebro, P. Bouvry, and L. Hogie, "A cellular MOGA for optimal broadcasting strategy in metropolitan MANETs," Comput. Commun., vol. 30, no. 4, pp. 685-697, 2007.\\[0pt]
[17] B. Dorronsoro, G. Danoy, P. Bouvry, and E. Alba, "Evaluation of different optimization techniques in the design of ad hoc injection networks," in Proc. Workshop Optim. Issues Grid Parallel Comput. Environ. Part HPCS, Nicossia, Cyprus, 2008, pp. 290-296.\\[0pt]
[18] H. Cheng and S. Yang, "Genetic algorithms with immigrant schemes for dynamic multicast problems in mobile ad hoc networks," Eng. Appl. Artif. Intell., vol. 23, no. 5, pp. 806-819, Aug. 2010.\\[0pt]
[19] H. Shokrani and S. Jabbehdari, "A novel ant-based QoS routing for mobile ad hoc networks," in Proc. 1st ICUFN, 2009, pp. 79-82.\\[0pt]
[20] C. Huang, Y. Chuang, and K. Hu, "Using particle swarm optimization for QoS in ad-hoc multicast," Eng. Appl. Artif. Intell., vol. 22, no. 8, pp. 11881193, Dec. 2009.\\[0pt]
[21] J. García-Nieto, J. Toutouh, and E. Alba, "Automatic tuning of communication protocols for vehicular ad hoc networks using metaheuristics," Eng. Appl. Artif. Intell., vol. 23, no. 5, pp. 795-805, Aug. 2010.\\[0pt]
[22] J. Kennedy and R. Eberhart, "Particle swarm optimization," in Proc. IEEE Int. Conf. Neural Netw., Nov. 1995, vol. 4, pp. 1942-1948.\\[0pt]
[23] K. V. Price, R. Storn, and J. Lampinen, Differential Evolution: A practical Approach to Global Optimization. London, U.K.: Springer-Verlag, 2005.\\[0pt]
[24] D. E. Goldberg, Genetic Algorithms in Search Optimization and Machine Learning. Reading, MA: Addison-Wesley, 1989.\\[0pt]
[25] S. Kirkpatrick, C. D. Gelatt, and M. P. Vecchi, "Optimization by simulated annealing," Science, vol. 220, no. 4598, pp. 671-680, May 1983.\\[0pt]
[26] The Network Simulator Project-Ns-2. [Online]. Available: \href{http://www.isi.edu/nsnam/ns/}{http://www.isi.edu/nsnam/ns/}\\[0pt]
[27] Y. Ge, T. Kunz, and L. Lamont, "Quality of service routing in ad-hoc networks using OLSR," in Proc. 36th HICSS, 2003, pp. 9-18.\\[0pt]
[28] A. Huhtonen, "Comparing AODV and OLSR routing protocols," in Proc. Telecommun. Softw. Multimedia, 2004, pp. 1-9.\\[0pt]
[29] Y. Huang, S. Bhatti, and D. Parker, "Tuning OLSR," in Proc. IEEE 17th Int. Symp. PIMRC, Helsiniki, Finland, 2006, pp. 1-5.\\[0pt]
[30] D. Nguyen and P. Minet, "Analysis of MPR selection in the OLSR protocol," in Proc. Int. Conf. Adv. Inf. Netw. Appl. Workshops, 2007, vol. 2, pp. 887-892.\\[0pt]
[31] E. Alba, S. Luna, and J. Toutouh, "Accuracy and efficiency in simulating VANETs," in Proc. MCO-Communications Computer Information Science, L. T. H. An, P. Bouvry, and T. P. Dinh, Eds., 2008, vol. 14, pp. 568-578.\\[0pt]
[32] D. Krajzewicz, M. Bonert, and P. Wagner, "The open source traffic simulation package SUMO," in Proc. RoboCup, Bremen, Germany, 2006, pp. 1-10.\\[0pt]
[33] E. Alba, G. Luque, J. García-Nieto, G. Ordonez, and G. Leguizamón, "MALLBA: A software library to design efficient optimisation algorithms," Int. J. Innov. Comput. Appl., vol. 1, no. 1, pp. 74-85, Apr. 2007.\\[0pt]
[34] D. J. Sheskin, Handbook of Parametric and Nonparametric Statistical Procedures. London, U.K.: Chapman \& Hall, 2007.\\[0pt]
[35] R. Eberhart and Y. Shi, "Comparing inertia weights and constriction factors in particle swarm optimization," in Proc. IEEE CEC, La Jolla, CA, 2000, vol. 1, pp. 84-88.\\[0pt]
[36] J. Toutouh and E. Alba, "An efficient routing protocol for green communications in vehicular ad-hoc networks," in Proc. 13th Annu. Conf. Companion GECCO, 2011, pp. 719-726.\\[0pt]
[37] V. Taliwal, D. Jiang, H. Mangold, C. Chen, and R. Sengupta, "Empirical determination of channel characteristics for DSRC vehicle-to-vehicle communication," in Proc. 1st ACM Int. Workshop VANET, 2004, p. 88.

\end{document}